\documentclass[nohyperref]{article}

\usepackage{booktabs}
\usepackage{cite}
\usepackage{amsmath,amssymb,amsfonts}
\usepackage{graphicx}

\usepackage{textcomp}
\usepackage{relsize}
\usepackage{xcolor}
\def\BibTeX{{\rm B\kern-.05em{\sc i\kern-.025em b}\kern-.08em
    T\kern-.1667em\lower.7ex\hbox{E}\kern-.125emX}}

\usepackage{tikz}

\usepackage[english]{babel}
\usepackage{blindtext}

\usepackage{float}
\usepackage{booktabs}

\usepackage{listings}

\usepackage[english]{babel}

\usepackage{hyperref}

\usepackage{balance}

\usepackage{xcolor}
\usepackage{colortbl}
\usepackage{amsmath}
\usepackage{pifont}
\usepackage{graphicx}
\usepackage{tabularx}
\usepackage[font=small,labelfont=bf]{caption}
\usepackage{adjustbox}
\usepackage[inline]{enumitem}
\usepackage{multirow}

\usepackage{microtype}
\usepackage{graphicx}
\usepackage{subfigure}
\usepackage{booktabs} 

\usepackage{amsmath}
\usepackage{amssymb}
\usepackage{mathtools}
\usepackage{amsthm}

\usepackage[capitalize,noabbrev]{cleveref}

\newcounter{MariosNOC}
\stepcounter{MariosNOC}

\newcounter{Alessandro}
\stepcounter{Alessandro}

\newcounter{Asterios}
\stepcounter{Asterios}

\newcounter{Ziyu}
\stepcounter{Ziyu}

\newcounter{RihanNOC}
\stepcounter{RihanNOC}

\usepackage{hyperref}

\usepackage[accepted]{icml2022}

\theoremstyle{plain}

\theoremstyle{definition}

\theoremstyle{remark}

\usepackage[textsize=tiny]{todonotes}

\newcommand{\thetitle}{Metadata Representations for Queryable ML Model Zoos}

\icmltitlerunning{\thetitle}
\begin{document}
\twocolumn[
\icmltitle{\thetitle}




\begin{icmlauthorlist}
\icmlauthor{Ziyu Li}{yyy}
\icmlauthor{Rihan Hai}{yyy}
\icmlauthor{Alessandro Bozzon}{yyy}
\icmlauthor{Asterios Katsifodimos}{yyy}
\end{icmlauthorlist}

\icmlaffiliation{yyy}{Delft University of Technology, The Netherlands}

\icmlcorrespondingauthor{}{z.li-14@tudelft.nl}

\icmlkeywords{Machine Learning, ICML}

\vskip 0.3in
]



\printAffiliationsAndNotice{} 

\begin{abstract}
Machine learning (ML) practitioners and organizations are building model zoos of pre-trained models, containing metadata describing properties of the ML models and datasets that are useful for reporting, auditing, reproducibility, and interpretability purposes. The metatada is currently not standardised; its expressivity is limited; and there is no interoperable way to store and query it. Consequently, model search, reuse, comparison, and composition are hindered. In this paper, we advocate for standardized ML model metadata representation and management, proposing a toolkit supported to help practitioners manage and query that metadata.

\end{abstract}

\section{Introduction}
To incentivize and enable the sharing and reuse of ML models, organizations are building repositories of pre-trained ML models, also known as \textit{model zoos} or \textit{model hubs}; known examples include HuggingFace \cite{huggingface}, Tensorflow Hub\cite{tensorflow}, and PyTorch Hub \cite{pytorch}. 
The metadata of a ML model can contain all the necessary information -- from a model's inference capabilities (e.g., identified object classes) to the architecture,  execution time, training dataset, hyper-parameter configuration, and evaluation performance. Metadata management are developed in the context of an ongoing effort to promote Trustworthy and Responsible AI\footnote{\url{https://partnershiponai.org/paper/responsible-publication-recommendations/}}. 

Metadata management for model zoos can facilitate MLOps and open up new opportunities for new and unexplored use cases. For instance:   
\begin{enumerate*}[label=\roman*)]
    \item retrieving models from large repositories with complex filtering conditions; 
    \item continuous integration of models in production;
    \item (semi-)automatic model composition; and
    \item advanced data management system.
\end{enumerate*}
We will further discuss use cases in \cref{sec:usecase}.

The potential of model zoos is currently hindered by the lack of a structured, queryable metadata format. Current repositories include a wide range of information, in a form of a model card \cite{mitchell2019model}, but such information is mostly for human consumption, making it hard for automatic extension or management. At the same time, the level of the detail remains coarse-grained: for instance, Amazon SageMaker, AzureML, MLflow \cite{zaharia2018accelerating} do not require mandatory reporting of the related metadata, except for model name and version. Practitioners have to search on external websites for further metadata information such as the data instances, and they even have to evaluate the model at hand in order to assess its performance. 

\cref{tab:query} lists some example queries that are practical for ML developers and users, and we will further discuss them in \cref{sec:usecase}. These queries require more fine-grained model metadata that current model repositories such as HuggingFace do not support. This highlights the need for detailed metadata of trained ML models and datasets in a structured and queryable representation.

As a step toward this goal, we present a metadata model to represent the metadata of ML models and related datasets. Based on this metadata model, we are currently developing a ML model metadata management platform (prototype available at \href{http://modelsearch.io}{modelsearch.io}) that can be used to query such metadata. 

The contributions of this paper go as follows:

\vspace{-4mm}
\begin{itemize}
\setlength\itemsep{0em}
        \item We showcase two use cases abstracted from  ML life-cycle, which advocates the need to define appropriate metadata representation for model zoos (\cref{sec:usecase}).  
        \item We propose a metadata model as the structured, queryable, and comprehensive metadata representation for model zoos (\cref{ssec:model}).
        \item We develop an early-stage tool to retrieve ML models and datasets based on our proposed metadata representation, which facilitates complex inference queries over model zoos (\cref{ssec:tool}).
    \end{itemize}
    
\vspace{-3mm}
\def\arraystretch{1.2}
\begin{table}[t]
    \centering
    \caption{Example queries}
    \label{tab:query}
    \vskip 0.15in
    \small
    \begin{tabular}{p{0.3\linewidth}|p{0.03\linewidth}p{0.55\linewidth}}
    \toprule
    Property  & ID &  Query \\
    \hline
    \multirow{6}{*}{Dataset information} & \multirow{3}{*}{1} &  Retrieve text classification models trained on dataset crowdsourced by at least a group of 50 people \\ \cline{2-3}
    & \multirow{3}{*}{2} & Find a dataset collected from COCO and OpenImage with all the images containing ``dog'' \\
    \hline
    \multirow{5}{*}{Model performance} & \multirow{3}{*}{3} &  Retrieve models trained on ImageNet with an accuracy higher than 90\% \\
    \cline{2-3}
    & \multirow{2}{*}{4} &  Which model performs the best on COCO for person detection task? \\
    \hline
    \multirow{4}{*}{Interpretability} & \multirow{2}{*}{5} &  Retrieve a person detection model with no gender bias  \\
    \cline{2-3}
    & \multirow{2}{*}{6} &  Retrieve text generation models that do not generate hate speech \\
    \hline
    
    \multirow{3}{*}{Hardware-related} & \multirow{3}{*}{7} &  Retrieve image classification models that are suitable to deploy on edge devices\\
    \hline
    \end{tabular}
\vskip -0.1in
\end{table}

\section{Use Cases}
\label{sec:usecase}

To demonstrate the usefulness of a formalized metadata representation,  in this section we explain two use cases with different stakeholders, where a comprehensive and well-structured  metadata representation is needed.

\subsection{Metadata Retrieval Throughout the ML Life-cycle}

Throughout the ML life-cycle, ML practitioners will require different metadata for tracking the ML model status, editing, comparing, or reporting.
An ML practitioner often needs to query models in large repositories with complex filtering conditions, e.g., data instance, performance, and inherit mechanism. 
In \cref{tab:query}, we list a few example queries revealing different properties of the metadata. For example, Query 1 and 2 require the metadata regarding the dataset, i.e., its attribute and source. Queries such as Query 3 to 6 require some other metadata properties. 
Query 7, on the other hand,  requires more complex information regarding the inference performance with specified hardware settings. For example, an edge device may have constraints such as limited computation power and storage. To answer this query, practitioners will need to obtain the model performance of different objectives, e.g., inference speed and memory footprint.

\subsection{(Semi-)Automatic Model Composition}
Now we introduce a more advanced yet common use case.
With metadata being captured and well-represented, ML practitioners can make good use of the models trained offline and apply them to answer complex, ad-hoc inference queries. 

\begin{figure}[t]
    \centering
  \includegraphics[width=0.5\textwidth]{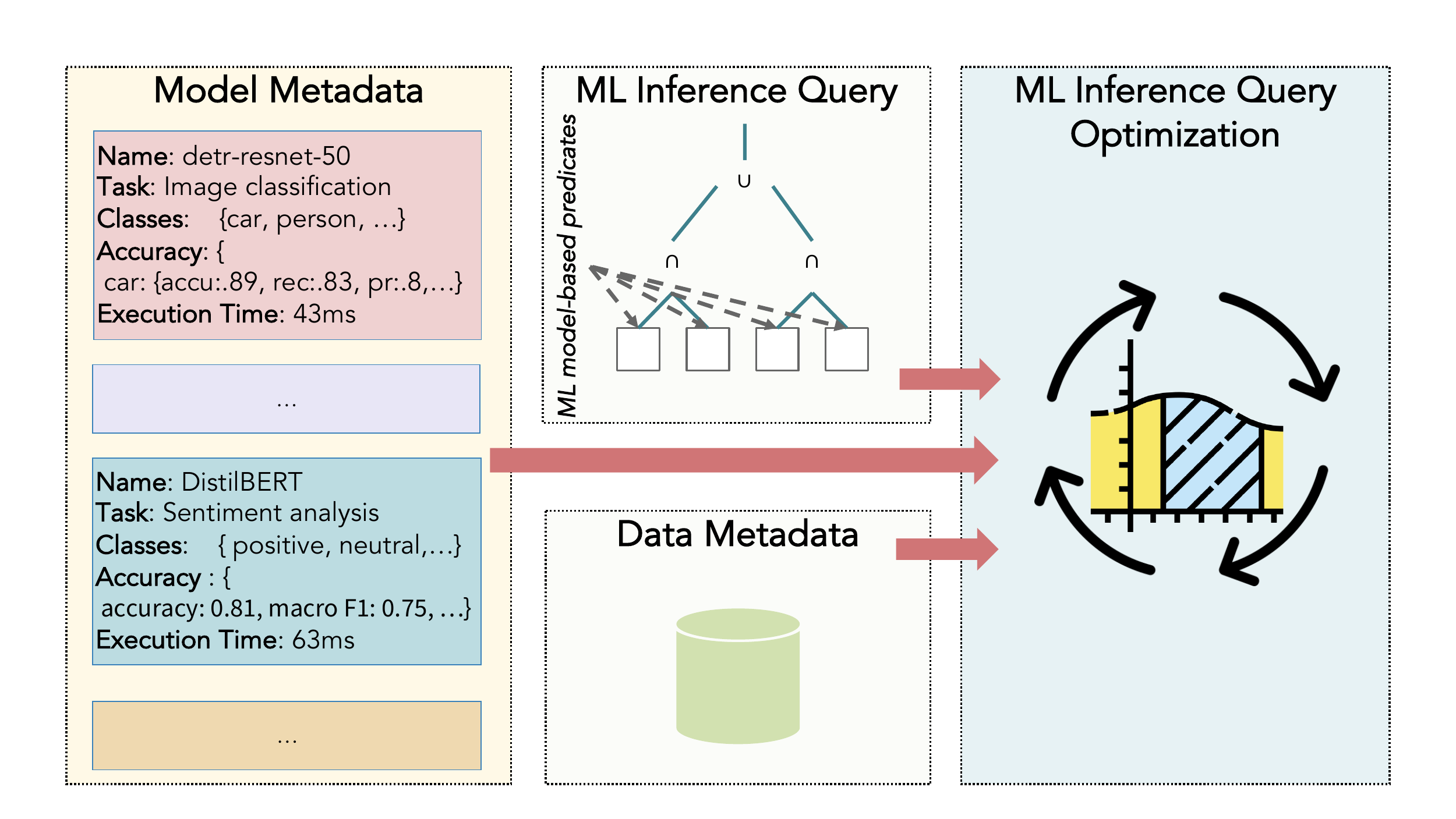}
  \caption{ML inference query optimization}
  \label{fig:optimizer}
\end{figure}

As shown in \cref{fig:optimizer}, the ad-hoc query can consist of multiple ML inference tasks with different dependencies and relations. Moreover, the query can be composed of specified constraints/requirements (e.g., latency and accuracy restrictions).
Practitioners can select a composition of models from the model zoo to answer the ad-hoc queries.
For example, an ML practitioner would like to design an application that can capture tweets with positive sentiment and a part-of-speech (PoS) tagger for data analysis. Since the data volume is significant and latency is also an essential factor to consider, the practitioner should select text classification model and PoS tagging model with fast inference speed. And the inference speed is greatly affected by the hardware being applied. If the application is deployed on the mobile phone, then the memory footprint is also a fundamental objective to be concerned with. 
To identify which set of models could best address the query and constraints, they may require information regarding the model performance with different objectives (e.g., accuracy, inference speed, memory footprint). The metadata of the dataset can also provide information to detect concept drift, for instance.

\vspace{5px}
State-of-the-art model repositories would not be able to support these use cases, as they  often maintain limited metadata, like the model name or the training dataset. A first step towards accessible repository is the specification of a rich and structured metadata format.

\section{Methodology}

This section defines a metadata model to represent different entities and relations. We also propose a practical tool to extract, store, and query the metadata. 
With our proposed comprehensive metadata in a structured representation and the tool, users can address the use cases discussed in the previous section by retrieving metadata in fine-grained details.


\subsection{Metadata Model}
\label{ssec:model}
\begin{figure}[t]
\begin{center}
  \includegraphics[width=0.51\textwidth]{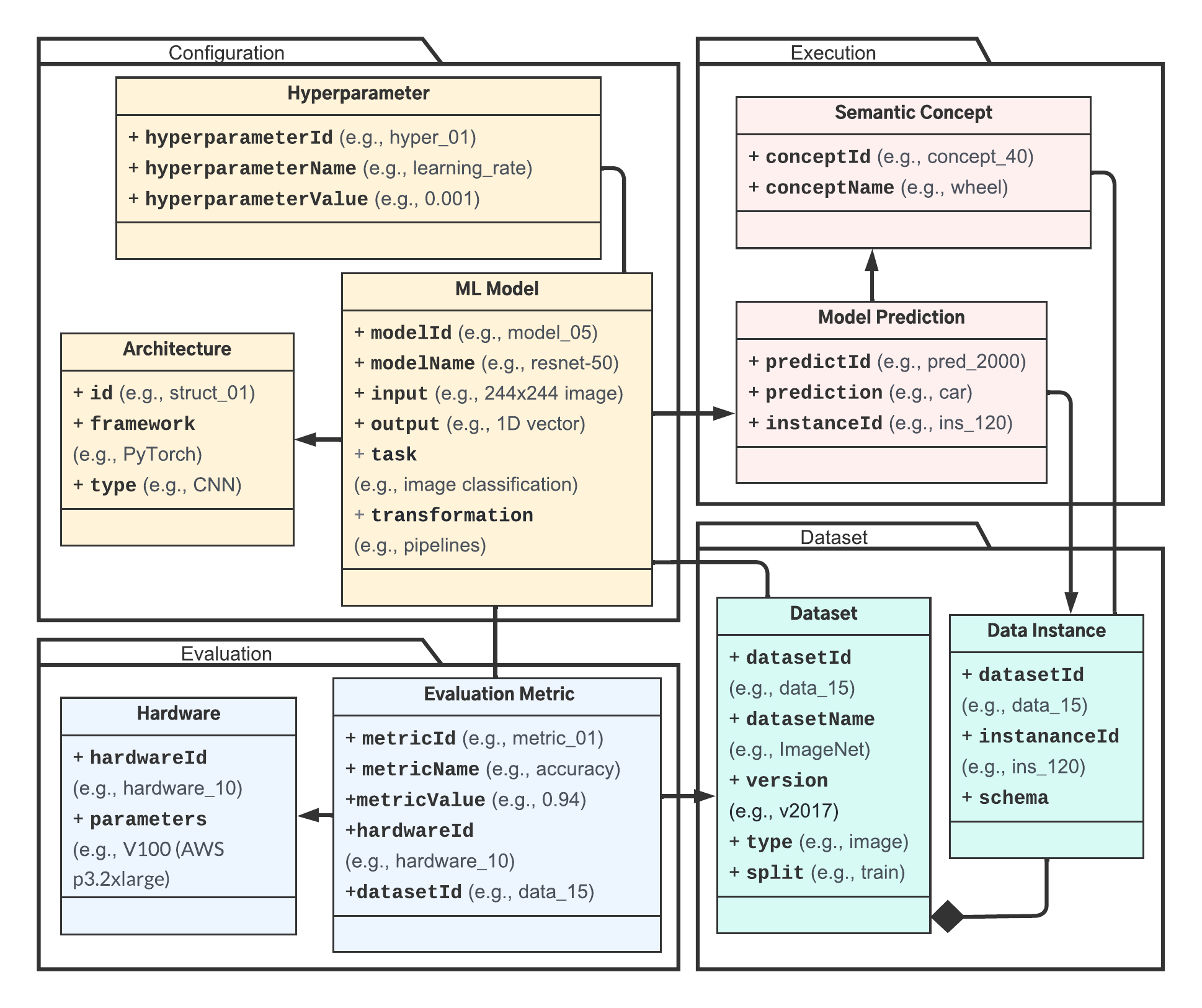}
  \caption{Modeling the metadata throughout ML life-cycle}
  \label{fig:model}
\end{center}
 \vskip -0.2in
\end{figure}

\cref{fig:model} presents a conceptual view of our proposed metadata model (from now on, \emph{meta-model}), described using UML class diagrams. The \emph{meta-model} composes of four packages: 
\begin{enumerate*}[label=\roman*)]
     \item the \texttt{Configuration} package, which defines the ML models associated with architecture, hyperparameter settings, input and output;
    \item the \texttt{Dataset} package, which represents information on datasets;
     \item the \texttt{Execution} package, which describes the inference results of the model, possibly enriched with description from a knowledge graph;
     \item the \texttt{Evaluation} package, which presents the run-time metrics obtained in specified hardware settings.
\end{enumerate*}

Due to space limitation, and for the sake of clarity, we present a compact version of the  \emph{meta-model}, which is amenable to implementation in different formats, including relational databases and Linked Data.

\subsubsection{Configuration package}

The package composes of three classes: \texttt{ML model}, \texttt{Hyperparameter}, and \texttt{Architecture}. 
The \texttt{ML Model} class encodes information about the \textit{input} and \textit{output} of the model, the prediction \textit{task}, and the data \textit{transformation} processes steps (e.g. data transformation and post-processing). For traditional ML model, the transformation pipelines can include steps for feature engineering. 
Such a recording of the ML model details enables exploration tasks such as model selection, data lineage, and visualization.  It also helps reproduce the ML model and compare it to others from different perspectives. The \texttt{Hyperparameter}, as the name suggests, capture the type and values of hyper-parameters used to train the specific model. Finally, the \texttt{Architecture} class contains information about the specific structural characteristic of a model, e.g. if it is a Feed-Forward Neural Network.

\subsubsection{Dataset Package}

The behaviour of a ML model heavily rely on the data it used. 
Thus the \emph{meta-model} includes a \texttt{Dataset} packaged, representing both  \texttt{Dataset}s and their \textit{Data Instance}s.
With the \texttt{Dataset} element, we present the metadata of the datasets that is significant for data management and reporting, e.g., data source, data version. 
With \texttt{Data Instance} we facilitate data provenance, for which it is of utmost importance to know which instances the model is trained on and how the data is utilized. 
Moreover, it is also significant to know if the data has revealed sensitive information, for example, violating the GDPR regulations. 



\subsubsection{Execution Package}
The \texttt{Execution} package includes two important classes. The \texttt{Model Prediction} class allows the description of inference metrics on one or more prediction output. Optionally a \texttt{Model Prediction} can be linked with a \emph{Semantic Concept} from a knowledge base, thus allowing complex reasoning. Consider the example that an ML practitioner is building an ML model for image classification of cars, and she tries to conduct model diagnosis. She may have several questions: what makes the model identify a car as a car? What are the semantic concepts that the model is capable to identify? Are the wheels that make it believe that it is a car? To support such use cases, the \emph{meta-model} allows to store information about inference performance on specific data instances, which can be used to describe the the behavior of a model, i.e., in what circumstances a model can perform well and why.  Such information and awareness of the model prediction, may significantly improve ML model interpretability in various applications such as health care, law enforcement and finance.

\subsubsection{Evaluation Package}

The performance evaluation of an ML model is critical during both the training and  deployment phases. The practitioners need to deploy a suitable ML model for the task given a specified environment, e.g., \texttt{Hardware} parameters. A mismatch of the deployment will lead to latency issues or reliability concerns, which results in user dissatisfaction. Thus a well-rounded evaluation of the model performance is significant. Hence, we have the \texttt{Hardware} preserving the hardware settings and \texttt{Evaluation Metric} with metric-related metadata. Different models will need different evaluation metrics, for instance, a regression model would have a \emph{Mean Square Error} - \textit{MSE}, while a multi-class classification model would have an \emph{accuracy} value for each class.

\begin{figure}[t]
  \centering
  \includegraphics[width=0.4\textwidth]{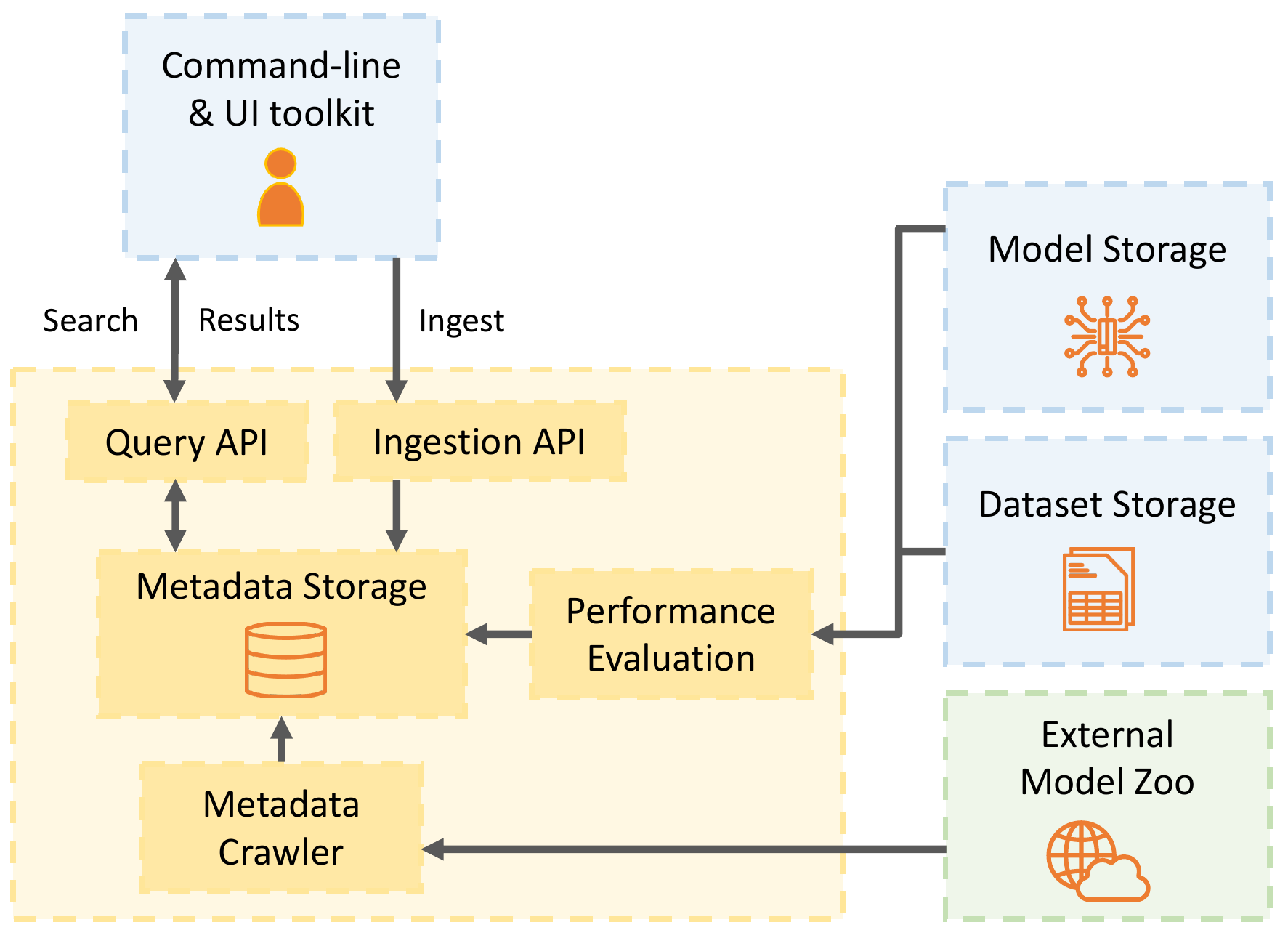}
  \caption{The structure of our metadata management tool}
  \label{fig:structure}
\end{figure}

\subsection{A Metadata Management Tool}
\label{ssec:tool}
We are currently developing a tool\footnote{\url{https://modelsearch.io}} to manage the metadata for model zoos, whose  structure is shown in \cref{fig:structure}. 

We collect metadata in the following three ways.
\begin{enumerate*}[label=(\roman*)]
    \item a user can add the metadata regarding a model or a dataset by filling in specified fields. Then such information is translated by the \emph{Ingestion API} into structured representation according to the above-mentioned metadata model, and stored in the \emph{Metadata Storage}. 
    \item  The tool can also automatically extract information from external model zoos, e.g., HuggingFace and PyTorch Hub. We can extract metadata from their API or information on the web page and record the source of the external metadata in our metadata field. 
    
    \item To gather the information regarding the model performance, we apply a third way to obtain the metadata. We obtain the performance evaluation by executing the model on a dataset with specified hardware settings. This process can be done by utilizing cloud resources.
\end{enumerate*}
 
By the time we are writing the paper, the tool is available to crawl information from the external model zoos and support querying the metadata stored in structured and queryable representations. Besides ingesting, extracting, and storing the metadata, our proposed tool allows a user to (i) retrieve the models that help them identify a model, (ii) compare multiple models, (iii) or explore the properties of models/data by composing queries on the metadata. 


\section{Related work}

Recent studies focus on different aspects of management during ML life-cycle, from model versioning, model reporting to model evaluation. Each is important for practitioners to manage and understand the models. We observe a gap among the profound works, a comprehensive and queryable metadata representation. With the metadata representation, we can thus better manage the ML models and data, including the interactions between them.

\emph{Metadata Captured in ML Versioning and Data Versioning}

Recent works developed tools/systems to manage ML models and datasets used in ML life-cycle. Works such as Modeldb \cite{vartak2016modeldb}, ModelHub \cite{miao2017towards}  and MLflow \cite{zaharia2018accelerating} developed tools to manage ML mdoels with different versions and captured the metadata of the models in different levels of details. However, they focused on the abstractions on the model, they lack information on the model performance under different hardware settings (e.g., inference speed, accuracy, memory footprint), and it paid little attention on the datasets that the models consume. 
For ML dataset management and versioning, research work such as Mldp \cite{agrawal2019data} and DataLab \cite{zhang2016datalab} can be complements to the above-mentioned model management system by supporting data management. 


\emph{Model Cards and Data Cards}

Recent research also focus on the reporting of models and dataset, covering aspects not only limited to basic informative components, but also including ethical, inclusive and fair considerations. Model cards \cite{mitchell2019model}, for example, proposed to include information regarding model intended use cases, potential pitfalls and other contexts that can improve model understanding. Similar idea also lies in data cards/sheets. Examples include \cite{gebru2021datasheets,miceli2021documenting,boyd2021datasheets}. Though model cards and data cards contain rich information, the Q\&A format is nonetheless unfriendly to machine to process and thus cannot be easily managed and retrieved. 

\emph{Model Performance Benchmarking}

A growing body of published work also focus into the benchmarking of ML model performance, such as MLperf \cite{mattson2020mlperf,reddi2020mlperf}, fathom \cite{adolf2016fathom}, Dawnbench \cite{coleman2017dawnbench}. These platforms covered a set of meatadata including metrics, training/inference configurations with specified hardware/software setting. Their focus is the report of the model performance at different ML life-cycle stages (training or inference). They paid little attention on the dataset the model used, whose path is provided as an argument filled by the user. The model process pipeline is also not covered besides the model scripts.


\section{Conclusion and outlook}
In this paper, we advocate for the need of a structured, queryable, and comprehensive metadata representation for model zoos. 
We propose a metadata model for such metadata representation to tackle different use cases. We also develop a tool that helps practitioners to manage and query the metadata.
Future work can develop tools/frameworks to extract more useful information from the textual description in the model/data cards by applying natural language processing techniques and mapping it to the predefined metadata representation.





\newpage
\nocite{langley00}

\bibliography{example_paper}
\bibliographystyle{icml2022}



\end{document}